\documentclass[letterpaper]{article} 
\usepackage[preprint]{aaai2027} 
\usepackage[hyphens]{url} 
\usepackage{graphicx} 
\usepackage{natbib} 
\usepackage{caption} 
\usepackage{booktabs}
\usepackage{multirow}
\usepackage[table]{xcolor}
\usepackage{placeins}
\usepackage{amsmath}
\usepackage{amssymb}
\usepackage{xcolor}
\title{Counterfactual Attention Policy Distillation for Temporal Video Grounding}
\author{
    Shaobo Ju$^{1}$,
    Haiyang Yu$^{2}$,
    Xuecheng Wu$^{3}$,
    Qiong Wu$^{1}$,
    Jiacong Wang$^{4}$,
    Fan Shi$^{2}$,\\
    Jun Peng$^{1}$,
    Yiyi Zhou$^{1}$\thanks{Corresponding Author.}
}
\affiliations{
    $^{1}$ Key Laboratory of Multimedia Trusted Perception and Efficient Computing,\\
    Ministry of Education of China, Xiamen University, 361005, P.R. China.\\
    $^{2}$ Fudan University.\\
    $^{3}$ MMLab, The Chinese University of Hong Kong.\\
    $^{4}$ University of the Chinese Academy of Sciences.\\
    jushaobo@stu.xmu.edu.cn
}

\begin{document}

\maketitle

\begin{abstract}
Temporal video grounding is a key capability of advanced \emph{Multimodal Large Language Models} (MLLMs) for the thorough understanding of video events, which is however often limited by repeated actions and visually similar contexts in long videos. 
In this paper, we study this issue from the perspective of \emph{On-policy distillation} (OPD) and propose a new training regime for MLLMs termed \emph{Counterfactual Attention Policy Distillation} (CAPD). 
In particular, OPD is a viable solution for MLLMs via providing dense teacher supervision on student-generated trajectories. 
But its next-token based teacher-student distillation is hard to identify the specific video segments supporting each predicted timestamp, which is critical for temporal grounding. 
In this case, CAPD measures how masking each temporal group changes the teacher's output distribution. 
The resulting counterfactual influence calibrates the teacher's attention and weights token-level distillation, allowing the student to learn the temporal evidence that affects boundary prediction.
To validate CAPD, we trained it on Qwen3-VL-8B-Instruct using only 2,500 samples for one epoch, and evaluated it on the TimeLens and multiple general video benchmarks. 
Experimental results show that CAPD improves average recall by 12.0\% relative to GRPO on TimeLens while preserving general video understanding, achieving comparable accuracy to the base model.
\end{abstract}

\section{Introduction}

\emph{Temporal video grounding} (TVG) aims to localize the video interval described by a language query. 
It is a fundamental capability for video retrieval, editing, and evidence-based question answering. 
Early studies formulate TVG as proposal ranking, span prediction, or set prediction~\cite{gao2017tall,zhang2020twodtan,mun2020lgi,zhang2020vslnet,lei2021momentdetr,li2022cvg}. 
Recent methods improve query-conditioned video representations, develop unified temporal localization frameworks, and leverage large-scale video-language pretraining~\cite{moon2023qddetr,lin2023univtg,bao2025vidgroup,yang2025timeexpert}. 
Despite this progress, accurately identifying precise event boundaries remains difficult when a short target event is surrounded by repeated actions, visually similar temporal context, or irrelevant but visually salient content.

\begin{figure}[t!]
    \centering
    \includegraphics[
        width=\columnwidth,
        trim=3cm 0.5cm 3cm 0cm,
        clip
    ]{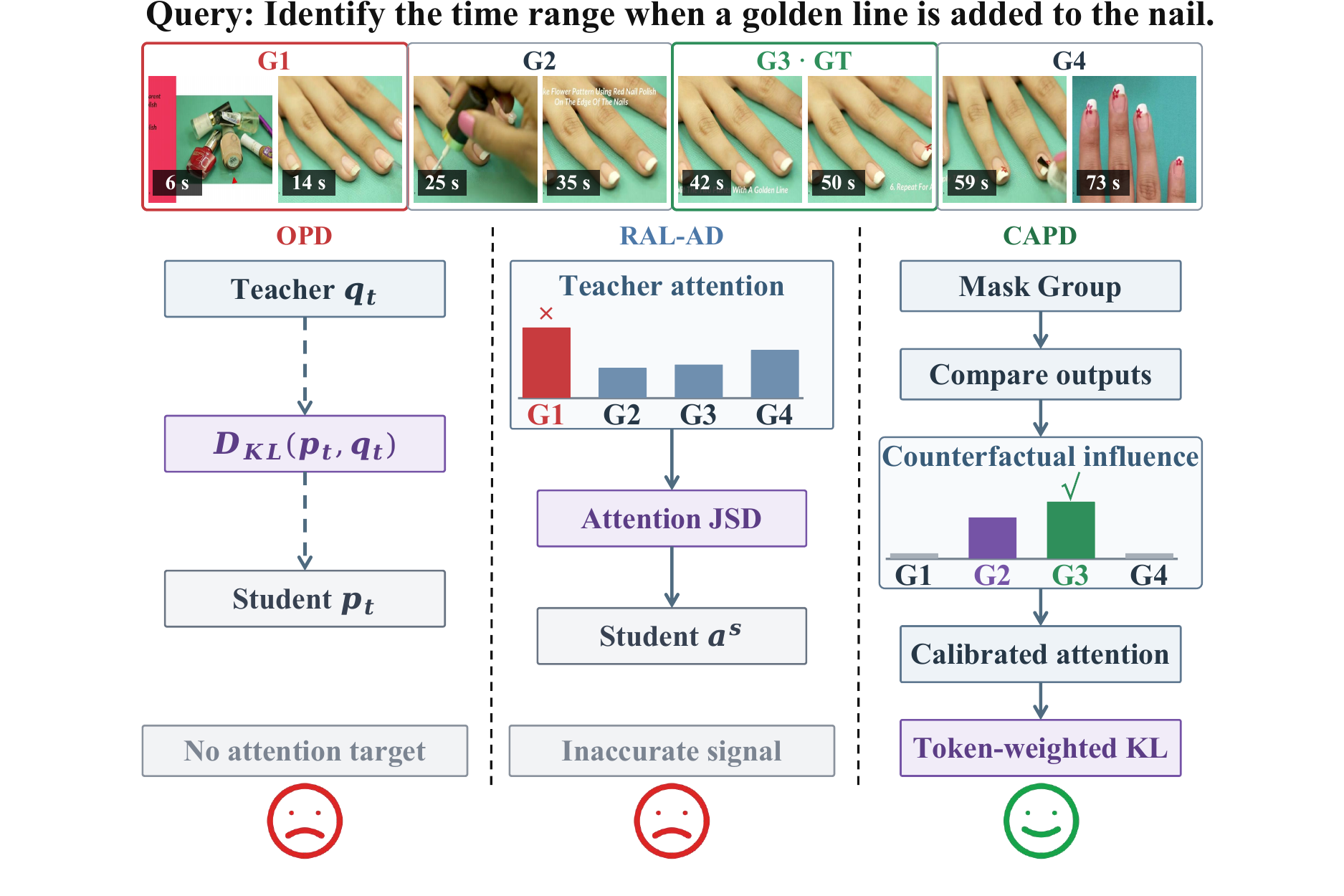}
    \caption{Comparison of OPD, RAL-AD, and CAPD. OPD only transfers output distributions, RAL-AD distills teacher attention, and CAPD uses counterfactual influence to calibrate attention and weight output-level distillation.}
    \label{fig:teaser}
\end{figure}

\emph{Multimodal large language models} (MLLMs)~\cite{bai2025qwen25vl,bai2025qwen3vl,ren2024timechat,huang2024vtimellm} formulate temporal video grounding as a text generation task, which directly generate the start and end timestamps from a video and a query without requiring a specially designed localization module~\cite{qian2024momentor,wang2024hawkeye}. 
Timestamp-aware encoding and instruction tuning improve temporal perception~\cite{ren2024timechat,huang2024vtimellm,qian2024momentor,wang2024hawkeye,yang2025timeexpert}, while recent post-training methods further strengthen grounding through verifiable rewards or dense teacher supervision~\cite{wang2025timer1,zhang2026timelens,li2026videoopd}. 
In particular, Li \emph{et al.} adopts \emph{On-Policy-Distillation} (OPD) ~\cite{li2026videoopd} to let the student generate its own on-policy trajectory and queries a stronger teacher on the same token prefixes. 
This setup provides the student with intensive monitoring signals, avoiding reliance solely on teacher-generated trajectories.

However, OPD only tells the student what to predict by matching the teacher's next-token distribution, but it doesn't provide guidance on which temporal regions should support each timestamp token. 
\emph{Reinforced Attention Learning} (RAL)~\cite{li2026ral} alleviates this gap by treating the teacher's internal attention distribution as an evidence-routing policy and transferring it to the student through on-policy attention distillation~\cite{li2026ral}. 
RAL provides an input-side learning signal that guides the student where to look in the video. 
However, where the teacher attends is not necessarily what its prediction depends on. 
As shown in Figure~\ref{fig:teaser}, masking a highly attended segment may leave the predicted timestamps unchanged, indicating that raw attention does not reliably identify decision-relevant evidence.

To address this limitation, we propose \textit{Counterfactual Attention Policy Distillation} (CAPD). As introduced above, OPD asks the student to match the teacher's output distribution, while RAL also asks it to match where the teacher attends. In particular, CAPD goes one step further by checking whether each attended video segment actually affects the teacher's prediction. Specifically, for each answer token, CAPD sums the teacher's attention within several contiguous temporal groups. It then masks one group at a time, runs the frozen teacher again under the same student-generated context, and then measures the change in its next-token distribution. A larger change indicates that the masked group contributes more to the prediction, so CAPD will increase the target attention on groups that cause larger output changes and reduces it on groups that cause little or no change. The student is trained to match both this corrected attention distribution and the teacher's output distribution, using the same on-policy rollouts as OPD. With these careful designs, CAPD can transfer decision-relevant temporal evidence while preserving OPD's dense output-level supervision, thereby improving temporal localization.

We validate CAPD with Qwen3-VL-8B using only 2,500 training examples for one epoch. 
Across three TimeLens-corrected benchmarks, CAPD achieves a 12.0\% improvement in average recall over GRPO. 
Importantly, these improvements in temporal understanding did not come at the expense of undermining general video understanding capabilities. 
CAPD remains comparable to the base model on several general video benchmarks, indicating that it largely retains the model's general capabilities after training.

Overall, our contributions are three-fold:
\begin{itemize}
    \item We identify the key shortcoming of existing OPD methods for temporal video grounding, \emph{i.e.}, OPD supervises outputs, while attention distillation shows where the teacher attends, not which regions determine its prediction.
    \item We propose a novel training paradigm for MLLMs termed Counterfactual Attention Policy Distillation (CAPD), which masks contiguous temporal groups to measure their effects on teacher predictions. The resulting influence calibrates teacher attention and reweights output-level distillation without altering the architecture or rollout.
    \item Under the same training settings, CAPD consistently outperforms OPD and RAL on  TimeLens while largely preserving general video understanding.
\end{itemize}

\begin{figure*}[t!]
    \centering
    \includegraphics[width=\textwidth, trim=0 4cm 0 0, clip]{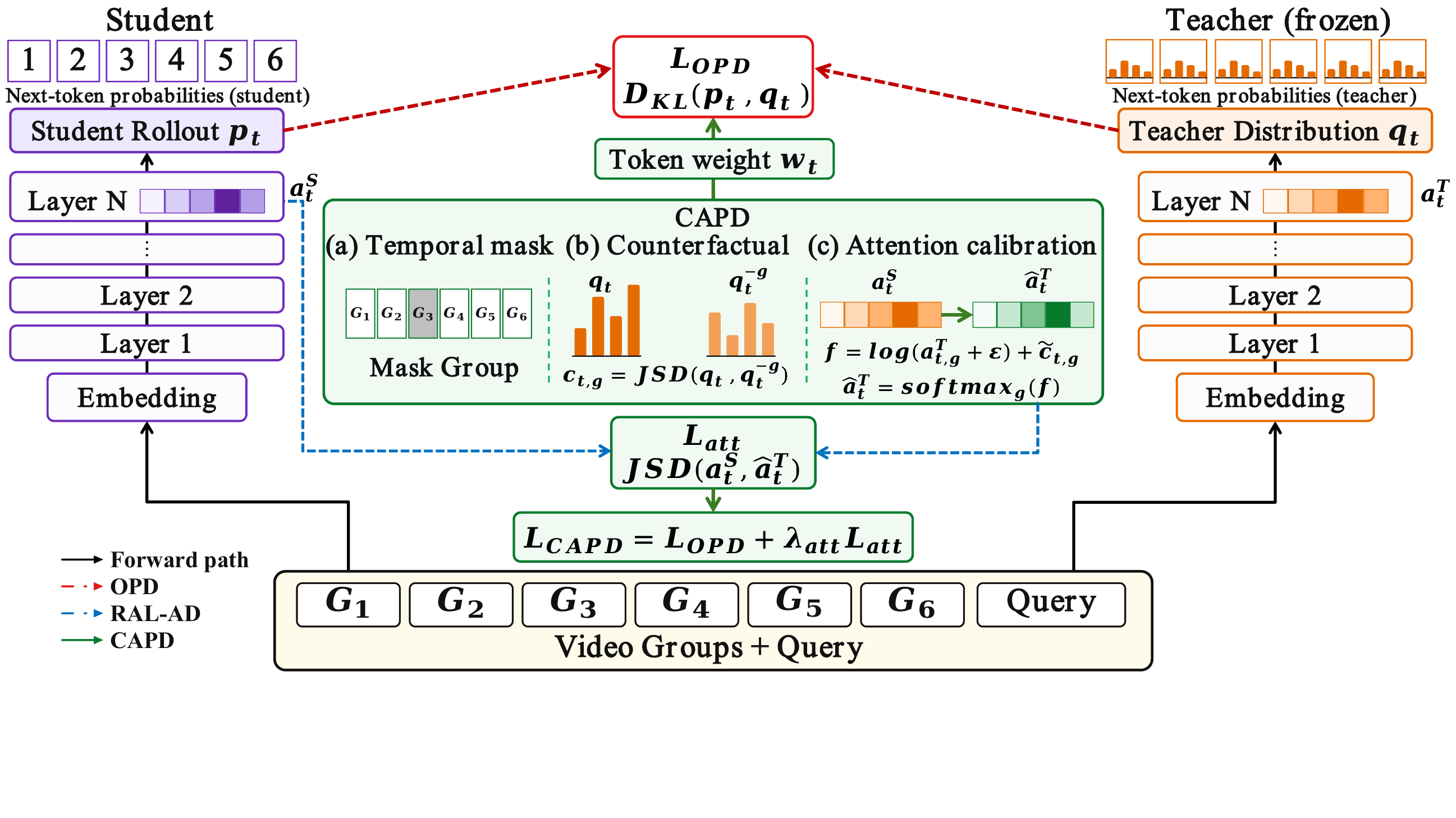}
    \caption{Illustration of the CAPD framework. Given the temporally grouped video tokens and query, the student generates an on-policy trajectory and the frozen teacher provides next-token supervision. CAPD masks each temporal group (a), estimates its counterfactual influence from the resulting teacher-output changes (b), and uses this influence to calibrate teacher attention and weight output-level distillation (c). The final objective combines output-level OPD with calibrated attention-policy learning.}
    \label{fig:method}
\end{figure*}

\section{Related Work}

\subsection{Temporal Video Grounding}

Temporal video grounding has evolved from specialized localization models to generative video MLLMs~\cite{wu2026survey}. Early approaches rank candidate segments, model pairwise moment relations, predict boundary spans, or perform set prediction~\cite{gao2017tall,zhang2020twodtan,zhang2020vslnet,lei2021momentdetr}. With the development of video MLLMs, TimeChat improves timestamp-aware long-video understanding through temporal instruction tuning, while VTimeLLM strengthens boundary perception using boundary-aware instruction data~\cite{ren2024timechat,huang2024vtimellm}. Recent methods further emphasize temporal reasoning. Time-R1 applies reinforcement learning, VTimeCoT introduces training-free visual temporal reasoning, and TAR constrains intermediate reasoning with progressively refined timestamp anchors~\cite{wang2025timer1,zhang2025vtimecot,guo2026tar}. TimeLens improves annotation quality and post-training for more reliable grounding, whereas Video-OPD provides dense teacher distributions on student-generated responses~\cite{zhang2026timelens,li2026videoopd}. Counterfactual learning has also been explored for temporal grounding: Zhai et al. synthesize altered video-query samples to mitigate moment bias, while Xu et al. construct counterfactual sequences for weakly supervised contrastive learning~\cite{zhai2022counterfactual,xu2025counterfactual,wang2025causalvtg}. Unlike these approaches, CAPD uses temporal-group interventions to measure a frozen teacher's token-level output sensitivity and transfers this decision-relevant evidence through on-policy attention distillation.

\subsection{On-Policy Distillation}

Recent surveys place on-policy distillation (OPD) within distribution-based knowledge distillation for large language models~\cite{yang2025kdsurvey,fang2026kdddsurvey,song2026opdsurvey}. Unlike off-policy distillation, which trains the student on fixed reference or teacher-generated prefixes, OPD uses trajectories generated by the current student and asks the teacher to supervise the states that the student actually visits, thereby reducing the gap between training and inference. 
MiniLLM~\cite{gu2024minillm} applies reverse KL to student-generated sequences to reduce exposure mismatch. GKD~\cite{agarwal2024gkd} generalizes OPD by mixing student-generated and fixed trajectories and supporting different divergence objectives. DistiLLM~\cite{ko2024distillm} combines skew KL with adaptive reuse of off-policy data to improve training efficiency and stability. Veto~\cite{jang2026veto} modifies the teacher target to reduce harmful updates on low-confidence tokens. Prefix OPD~\cite{zhang2026prefixopd} supervises only useful reasoning prefixes and stops rollouts early to reduce training cost. TIP~\cite{xu2026tip} estimates token importance from student uncertainty and teacher--student disagreement. Self-Distilled Reasoner~\cite{zhao2026selfdistilled} uses the same model as a teacher with access to extra information and as a student with only the original input, avoiding the need for a separate teacher model. VOLD~\cite{bousselham2026vold} transfers reasoning from a text-only language model to a vision--language model through reinforcement learning and OPD. Video-OPD~\cite{li2026videoopd} applies OPD to temporal video grounding. RAL~\cite{li2026ral} treats attention distributions as latent policies and distills them on student-generated trajectories. Although attention can be an effective distillation target~\cite{li2024attentiontransfer}, it may not faithfully show which inputs determine the model prediction~\cite{jain2019attention,wiegreffe2019attention}. CAPD therefore measures the counterfactual influence of temporal groups and uses it to calibrate teacher attention before transferring it to the student.

\section{Method}

\subsection{Overview}

In this paper, we propose CAPD, a post-training method for temporal video grounding, whose overall framework is illustrated in Figure~\ref{fig:method}. CAPD aims to transfer not only the teacher's output distribution but also the temporal evidence that directly affects its prediction. Standard OPD~\cite{li2026videoopd} aligns the student with the teacher's next-token distributions, providing dense supervision for timestamp generation but no explicit guidance about which temporal regions should be used for prediction. To address this gap, RAL~\cite{li2026ral} treats attention distributions as policies, revealing where the teacher attends. However, a highly attended region may have little effect on the prediction. CAPD addresses this limitation by using counterfactual interventions to identify decision-relevant temporal evidence.

For each video-query pair, the student first generates an on-policy response with predicted timestamps, and the frozen teacher evaluates the same student-generated prefixes to provide next-token distributions and output-to-input attention. The former supervises the output, while the latter indicates where the teacher attends.

CAPD consists of five steps. We first perform on-policy distillation and construct a temporal-group attention policy by aggregating attention over temporal groups, non-visual prompt tokens, and previous response tokens. We then estimate counterfactual group importance by masking each temporal group and measuring the resulting change in the teacher's output distribution. We use this importance for counterfactual attention calibration toward influential groups and counterfactual token weighting for outputs that depend more strongly on visual evidence. Finally, we combine the weighted OPD loss with the calibrated attention loss.

\subsection{On-Policy Distillation}

Given a video $V$ and a text query $q$, we denote the trainable student by $\pi_{\mathrm{s}}(\cdot;\theta)$ and the frozen teacher by $\pi_{\mathrm{t}}(\cdot;\phi)$. The response sampled from the student is 
\begin{equation}
 y=(y_1,\ldots,y_L)\sim\pi_{\mathrm{s}}(\cdot\mid V,q),
 \label{eq:trajectory}
\end{equation}
where $L$ is the number of response tokens included in distillation and $i$
indexes an output position. At position $i$, both models receive the same
student-generated prefix $y_{<i}$. For either model
$m\in\{\mathrm{s},\mathrm{t}\}$, its next-token distribution is
\begin{equation}
 p_{m,i}=\pi_m(\cdot\mid V,q,y_{<i}).
 \label{eq:token_distribution}
\end{equation}
We define the output-level discrepancy at this position as
\begin{equation}
 d_i=\mathrm{KL}\!\left(p_{\mathrm{s},i}\,\|\,p_{\mathrm{t},i}\right).
 \label{eq:token_kl}
\end{equation}
OPD averages this discrepancy over the sampled response:
\begin{equation}
 \mathcal{L}_{\mathrm{OPD}}=\frac{1}{L}\sum_{i=1}^{L}d_i.
 \label{eq:opd}
\end{equation}
This objective tells the student which output distribution to match, but it
does not tell the student which part of the video supports each output token.

\subsection{Temporal-Group Attention Policy}

First, we divide the video into \(G\) temporal groups, assigning all visual tokens corresponding to the same frame to the same group. 
At the final Transformer layer, let $A_{m,i,j}$ denote model $m$'s self-attention probability from output position $i$ to input position $j$, after applying the same positional encoding and masks as in the forward and averaging over attention heads. 
Let $\mathcal{I}_g$ contain the indices of all visual tokens in temporal group $g$. The attention mass assigned to that group is
\begin{equation}
 a_{m,i,g}=\sum_{j\in\mathcal{I}_g}A_{m,i,j}.
 \label{eq:group_attention}
\end{equation}
In addition to the $G$ temporal groups, we separately calculate attention to non-visual prompt tokens and previously generated response tokens. 
The resulting attention policy is
\begin{equation}
 \mathbf{a}_{m,i}=
 [a_{m,i,1},\ldots,a_{m,i,G},a_{m,i,\mathrm{pr}},a_{m,i,\mathrm{pf}}]
 \in\Delta^{G+2}.
 \label{eq:policy}
\end{equation}
These two language-token groups preserve how much the token depends on the prompt and its response prefix. 
Otherwise, the visual attention would be incorrectly renormalized to one. 
In our RAL-AD baseline, attention is directly used for distillation, achieved by minimizing the Jensen--Shannon divergence between $\mathbf{a}_{\mathrm{s},i}$ and $\mathbf{a}_{\mathrm{t},i}$ together with the OPD loss. 
It transfers where the teacher looks, but it does not establish whether a visual group changes the teacher's prediction.

\subsection{Counterfactual Group Importance}

To measure the effect of temporal group $g$, we construct a counterfactual video $V^{(-g)}$. 
For every visual token in the group, its feature is replaced by the mean feature at the same spatial location over the remaining temporal positions. 
This removes the group's temporal content while preserving the input shape, spatial layout, and positional indices. 
The frozen teacher then evaluates $V^{(-g)}$ using the same query and the same student prefix $y_{<i}$ as in the full-video forward pass. 
We denote the resulting next-token distribution by $p_{\mathrm{t},i}^{(-g)}$.
The importance of group $g$ to output position $i$ is the change in the teacher's next-token distribution after this removal:
\begin{equation}
 c_{i,g}=\mathrm{JSD}\!\left(
 p_{\mathrm{t},i},p_{\mathrm{t},i}^{(-g)}
 \right).
 \label{eq:credit}
\end{equation}
Thus, $a_{\mathrm{t},i,g}$ and $c_{i,g}$ answer different questions. 
The former measures how much attention the teacher routes to group $g$, whereas the latter measures whether that group affects the teacher's output. 
We summarize the total visual dependence of token $i$ as
\begin{equation}
 u_i=\sum_{g=1}^{G}c_{i,g}.
 \label{eq:total_credit}
\end{equation}
Since the absolute scale of $u_i$ varies across responses, we use a bounded gate
\begin{equation}
 \rho_i=\frac{u_i}{u_i+s},
 \label{eq:credit_gate}
\end{equation}
where $s$ is the median of the positive $u_i$ values within the current response. 
If no token has positive importance, we set every $\rho_i$ to zero. 
This gate prevents numerically small counterfactual changes from causing a large adjustment.

\begin{table*}[t]
\centering
\caption{Comparison of CAPD with proprietary models, open-source models, and post-training frameworks, following Video-OPD~\cite{li2026videoopd}. 
\textbf{Bold} and \underline{underline} indicate the best and second-best results among the post-training methods.}
\label{tab:main_results}
\begingroup
\small
\renewcommand{\arraystretch}{0.94}
\setlength{\tabcolsep}{2.0pt}
\begin{tabular*}{\textwidth}{@{\extracolsep{\fill}}lcccccccccc@{}}
\toprule
\multicolumn{1}{c}{\multirow[c]{2}{*}{\textbf{Method}}}
& \multicolumn{3}{c}{\textbf{Charades-TimeLens}}
& \multicolumn{3}{c}{\textbf{ActivityNet-TimeLens}}
& \multicolumn{3}{c}{\textbf{QVHighlights-TimeLens}} & \multirow[c]{2}{*}{\textbf{Avg.}} \\
\cmidrule(lr){2-4}\cmidrule(lr){5-7}\cmidrule(lr){8-10}
& \textbf{R@0.3} & \textbf{R@0.5} & \textbf{R@0.7}
& \textbf{R@0.3} & \textbf{R@0.5} & \textbf{R@0.7}
& \textbf{R@0.3} & \textbf{R@0.5} & \textbf{R@0.7} & \\
\midrule
\rowcolor{gray!15}\multicolumn{11}{l}{\textit{Proprietary Models}} \\
\textcolor{gray!90}{GPT-4o             } & \textcolor{gray!90}{60.6} & \textcolor{gray!90}{44.5} & \textcolor{gray!90}{23.5} & \textcolor{gray!90}{55.2} & \textcolor{gray!90}{41.4} & \textcolor{gray!90}{25.8} & \textcolor{gray!90}{69.0} & \textcolor{gray!90}{54.8} & \textcolor{gray!90}{38.5} & \textcolor{gray!90}{45.9} \\
\textcolor{gray!90}{GPT-5              } & \textcolor{gray!90}{59.3} & \textcolor{gray!90}{42.0} & \textcolor{gray!90}{22.0} & \textcolor{gray!90}{57.4} & \textcolor{gray!90}{44.9} & \textcolor{gray!90}{30.4} & \textcolor{gray!90}{72.4} & \textcolor{gray!90}{60.4} & \textcolor{gray!90}{46.4} & \textcolor{gray!90}{48.4} \\
\textcolor{gray!90}{Gemini-2.0-Flash   } & \textcolor{gray!90}{66.4} & \textcolor{gray!90}{53.5} & \textcolor{gray!90}{27.1} & \textcolor{gray!90}{62.9} & \textcolor{gray!90}{54.0} & \textcolor{gray!90}{37.7} & \textcolor{gray!90}{76.2} & \textcolor{gray!90}{66.4} & \textcolor{gray!90}{48.3} & \textcolor{gray!90}{54.7} \\
\textcolor{gray!90}{Gemini-2.5-Flash   } & \textcolor{gray!90}{68.7} & \textcolor{gray!90}{56.1} & \textcolor{gray!90}{30.6} & \textcolor{gray!90}{66.8} & \textcolor{gray!90}{57.5} & \textcolor{gray!90}{41.3} & \textcolor{gray!90}{78.2} & \textcolor{gray!90}{69.4} & \textcolor{gray!90}{55.0} & \textcolor{gray!90}{58.2} \\
\textcolor{gray!90}{Gemini-2.5-Pro     } & \textcolor{gray!90}{74.1} & \textcolor{gray!90}{61.1} & \textcolor{gray!90}{34.0} & \textcolor{gray!90}{72.3} & \textcolor{gray!90}{64.2} & \textcolor{gray!90}{47.1} & \textcolor{gray!90}{84.1} & \textcolor{gray!90}{75.9} & \textcolor{gray!90}{61.1} & \textcolor{gray!90}{63.8} \\
\midrule
\rowcolor{gray!15}\multicolumn{11}{l}{\textit{Open-Source Models}} \\
VideoChat-Flash-7B~\cite{li2026videochatflash}  & 60.2 & 37.9 & 17.8 & 35.5 & 21.8 & 10.5 & 45.2 & 30.6 & 16.7 & 30.7 \\
VideoChat-R1-7B~\cite{li2025videochatr1}     & 51.9 & 30.8 & 11.7 & 35.0 & 23.9 & 11.3 & 29.3 & 19.1 & 9.4  & 24.7 \\
Time-R1-7B~\cite{wang2025timer1}          & 57.9 & 32.0 & 16.9 & 44.8 & 31.0 & 19.0 & 65.8 & 51.5 & 36.1 & 39.4 \\
TVG-R1-7B~\cite{chen2025tvg}       & 44.5 & 23.6 & 12.4 & 46.7 & 31.0 & 18.6 & 55.8 & 41.2 & 28.0 & 33.5 \\
VideoChat-R1.5-7B~\cite{yan2025videochatr15} & 46.4 & 24.0 & 10.4 & 40.6 & 25.3 & 16.4 & 62.2 & 44.5 & 28.3 & 33.1 \\
MiMo-VL-7B~\cite{xiaomi2025mimovl}          & 57.9 & 42.6 & 20.5 & 49.3 & 38.7 & 22.4 & 57.1 & 42.6 & 28.4 & 39.9 \\
Qwen2.5-VL-7B~\cite{bai2025qwen25vl}   & 58.1 & 35.1 & 18.2 & 47.2 & 32.5 & 20.2 & 55.0 & 41.7 & 29.3 & 37.5 \\
Qwen3-VL-8B-Instruct~\cite{bai2025qwen3vl} & 61.7 & 41.5 & 23.1 & 41.2 & 30.7 & 20.0 & 46.6 & 38.2 & 29.5 & 36.9 \\
\midrule
\rowcolor{gray!15}\multicolumn{11}{l}{\textit{Post-Training Frameworks (Based on Qwen3-VL-8B-Instruct)}} \\
GRPO~\cite{shao2024deepseekmath}   & 72.7 & 44.4 & 27.6 & 58.6 & 42.7 & 32.1 & 69.8 & 53.0 & 41.5 & 49.2 \\
Video-OPD~\cite{li2026videoopd} & 73.1 & 45.8 & \textbf{32.4} & 60.5 & 45.6 & 35.8 & \underline{73.8} & \underline{60.3} & \underline{50.4} & 53.1 \\
Vanilla OPD~\cite{song2026opdsurvey} & 75.5 & 48.5 & 30.7 & 62.6 & \underline{46.6} & 36.0 & 71.8 & 57.2 & 46.1 & 52.8 \\
RAL-AD~\cite{li2026ral}      & \underline{75.6} & \underline{48.6} & 31.3 & \underline{62.9} & \underline{46.6} & \underline{36.3} & 72.6 & 58.1 & 47.1 & \underline{53.3} \\
\textbf{CAPD (Ours)}        & \textbf{76.1} & \textbf{49.8} & \underline{32.0} & \textbf{64.9} & \textbf{48.8} & \textbf{38.3} & \textbf{74.4} & \textbf{60.9} & \textbf{50.7} & \textbf{55.1} \\
\bottomrule
\end{tabular*}
\endgroup
\end{table*}

\subsection{Counterfactual Attention Calibration}

For each output position $i$, we standardize the $G$ group scores $c_{i,1:G}$, obtaining $\widetilde{\mathbf{c}}_i$. 
We then modify the teacher's visual attention logits according to counterfactual importance:
\begin{equation}
 z_{i,b}=
 \begin{cases}
 \log(a_{\mathrm{t},i,b}+\epsilon)
 +\rho_i\widetilde c_{i,b},
 & b\in\{1,\ldots,G\},\\
 \log(a_{\mathrm{t},i,b}+\epsilon),
 & b\in\{\mathrm{pr},\mathrm{pf}\},
 \end{cases}
 \label{eq:calibration_logits}
\end{equation}
where $\epsilon$ is a small constant for numerical stability. 
The calibrated teacher policy is
\begin{equation}
 \widehat{\mathbf{a}}_{\mathrm{t},i}=\mathrm{softmax}(\mathbf{z}_i).
 \label{eq:calibrated_policy}
\end{equation}
A temporal group receives more target mass when its removal changes the teacher's prediction more than the other groups for the same token, and less target mass when its effect is below average. 
When the teacher shows little total visual dependence, $\rho_i$ keeps the calibrated policy close to the original attention policy.

We train the student to match this target using
\begin{equation}
 \mathcal{L}_{\mathrm{att}}=
 \frac{\sum_{i=1}^{L}\gamma_i\,
 \mathrm{JSD}(\mathbf{a}_{\mathrm{s},i},
 \widehat{\mathbf{a}}_{\mathrm{t},i})}
 {\sum_{i=1}^{L}\gamma_i+\epsilon},
 \label{eq:attention_loss}
\end{equation}
where $\gamma_i=\rho_i u_i$ assigns more attention supervision to tokens with obvious visual dependence. 

\subsection{Counterfactual Token Weighting}

The same counterfactual signal is also used to emphasize output tokens whose predictions depend more strongly on the video. 
We first normalize each $u_i$ by the mean visual dependence within the response:
\begin{equation}
 \overline u_i=
 \frac{u_i}{L^{-1}\sum_{j=1}^{L}u_j+\epsilon}.
 \label{eq:normalized_credit}
\end{equation}
We convert this value into a capped preliminary weight:
\begin{equation}
 \widetilde w_i=\min(1+\overline u_i,w_{\max}).
 \label{eq:preliminary_weight}
\end{equation}
Finally, we divide all preliminary weights by their response mean so that the average weight remains one:
\begin{equation}
 w_i=\frac{\widetilde w_i}
 {L^{-1}\sum_{j=1}^{L}\widetilde w_j}.
 \label{eq:token_weight}
\end{equation}
The weighted output-level loss is
\begin{equation}
 \mathcal{L}_{\mathrm{OPD}}^{\mathrm{w}}=
 \frac{\sum_{i=1}^{L}w_i d_i}{\sum_{i=1}^{L}w_i}.
 \label{eq:weighted_opd}
\end{equation}
CAPD combines output-distribution matching with calibrated attention-policy matching:
\begin{equation}
 \mathcal{L}_{\mathrm{CAPD}}=
 \mathcal{L}_{\mathrm{OPD}}^{\mathrm{w}}
 +\lambda_{\mathrm{att}}\mathcal{L}_{\mathrm{att}}.
 \label{eq:total}
\end{equation}

\section{Experiments}

\subsection{Implementation Details}

We initialize the student from Qwen3-VL-8B-Instruct and use the GRPO-post-trained Qwen3-VL-32B model as the frozen teacher. 
All methods are trained for one epoch on the 2,500 examples provided by Video-OPD~\cite{li2026videoopd}. 
Videos are sampled at 2 FPS under an 8,192 visual token budget. 
Training uses full-parameter optimization with a learning rate of $1\times10^{-6}$, and DeepSpeed ZeRO-3 on 8 $\times$ H20 GPUs.

For CAPD, we use $G=8$ temporal groups and extract student and teacher policies from the last language-model layer. 
We use fused AdamW with a learning rate of $1\times10^{-6}$, set the random seed to 42, the attention-loss coefficient $\lambda_{\mathrm{att}}=0.25$, and the maximum token weight $w_{\max}=5$.
RAL-AD uses the on-policy attention-distillation variant introduced by RAL, which shares the same training configuration, directly distills teacher attention. 
Vanilla OPD disables both attention distillation and counterfactual token weighting. 
All experiments use the same training examples, configuration, and prompts.

\subsection{Benchmarks and Metrics}

We evaluate temporal grounding on Charades-TimeLens, ActivityNet-TimeLens, and QVHighlights-TimeLens, which use the corrected temporal annotations introduced by TimeLens~\cite{zhang2026timelens}. 
The evaluation sets contain 3,363, 4,500, and 1,541 queries. For a predicted interval $P$ and ground-truth interval $G$, temporal IoU is $|P\cap G|/|P\cup G|$. We report Recall at IoU thresholds 0.3, 0.5, and 0.7.

We additionally evaluate Video-MME~\cite{fu2025videomme}, LongVideoBench~\cite{wu2024longvideobench}, and LVBench~\cite{wang2025lvbench} to measure whether temporal post-training preserves general video understanding. 
Video-MME is a benchmark spanning short, medium and long videos. LongVideoBench contains 3,763 videos in four duration ranges from 8 seconds to 60 minutes. LVBench contains 103 videos, with an average video length of 68 minutes.

\subsection{Quantitative Analysis}

\subsubsection{Performance Comparison with Other Methods}
We first compare CAPD with existing video MLLMs and temporal-grounding post-training methods on three TimeLens-corrected benchmarks~\cite{zhang2026timelens} in Table~\ref{tab:main_results}. The results of Video-OPD~\cite{li2026videoopd} are included as reference, while Vanilla OPD, RAL-AD, and CAPD are trained for one epoch and evaluated under the same settings. From these results, we can first observe that post-training substantially improves the temporal grounding ability of video MLLMs. However, the improvements of existing methods vary across datasets and IoU thresholds, indicating that retrieving a relevant segment does not necessarily produce accurate temporal boundaries. Compared with Vanilla OPD, RAL-AD achieves better performance by aligning the raw teacher attention. More importantly, CAPD reaches an average of 55.1, improving over GRPO by 5.9 points (12.0\% relative), and achieves the best average performance among the compared post-training methods. It also improves most recall metrics, especially under the stricter IoU threshold. These results support the effectiveness of counterfactual calibration for learning decision-relevant temporal evidence. Across the three benchmarks, the gains are not concentrated in a single dataset. The larger improvements at stricter IoU thresholds indicate that CAPD not only retrieves an overlapping moment, but also refines the temporal extent of the prediction. This suggests that counterfactual calibration helps the model focus on the evidence needed to identify the queried event and determine its boundaries.

We evaluate the general video understanding capability of CAPD, Vanilla OPD~\cite{song2026opdsurvey}, RAL-AD~\cite{li2026ral}, and the base model on Video-MME~\cite{fu2025videomme}, LongVideoBench~\cite{wu2024longvideobench}, and LVBench~\cite{wang2025lvbench}, as reported in Table~\ref{tab:general_video}. These benchmarks cover a broad range of general video understanding tasks, providing a complementary evaluation to the temporal grounding benchmarks above. CAPD maintains comparable performance to Qwen3-VL-8B~\cite{bai2025qwen3vl} and the baselines across all three benchmarks, indicating that the counterfactual distillation objective does not interfere with the model's general video understanding capability. This result suggests that CAPD's training signal is sufficiently targeted to temporal grounding, leaving the broader video comprehension abilities of the base model intact.

\paragraph{Training Efficiency}
Under the same settings, CAPD requires 10.2 hours of training compared with 6.4 hours for OPD, corresponding to a $1.60\times$ training cost. The additional computation mainly comes from the teacher forward passes used for counterfactual temporal analysis. Nevertheless, the peak memory on GPU increases by only 5.21 GiB, allowing CAPD to remain trainable under the same setting as OPD.

\begin{table}[t]
\centering
\caption{Comparison of Qwen3-VL-8B-Instruct and its post-trained variants on general video understanding benchmarks.}
\label{tab:general_video}
\setlength{\tabcolsep}{4.5pt}
\resizebox{\columnwidth}{!}{%
\begin{tabular}{lccc}
\toprule
Method & Video-MME & LongVideoBench & LVBench \\
\midrule
Qwen3-VL-8B-Instruct  & 70.8 & 64.0 & 53.0 \\
$+$ Vanilla OPD       & 71.1 & 64.0 & 52.6 \\
$+$ RAL-AD            & 71.2 & 63.8 & 52.4 \\
$+$ CAPD              & 71.1 & 64.0 & 52.5 \\
\bottomrule
\end{tabular}}
\end{table}

\subsubsection{Ablation Study}

\begin{figure}[t!]
    \centering
    \includegraphics[
        width=\columnwidth,
        trim=0cm 0cm 0cm 0cm,
        clip
    ]{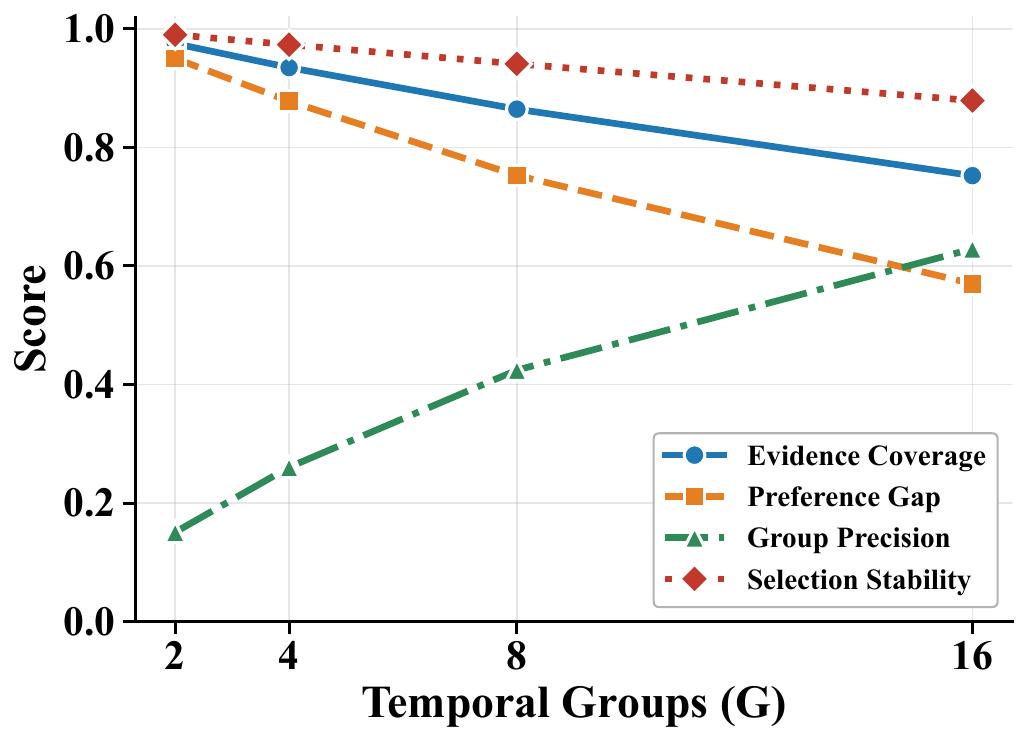}
    \caption{Analysis of evidence coverage, preference gap, group precision, and selection stability under different numbers of temporal groups.}
    \label{fig:ablation_1}
\end{figure}

\begin{table}[t]
\centering
\caption{Component ablation on TimeLens. C-TL, A-TL, and QV-TL are the means of R@0.3, R@0.5, and R@0.7 on each benchmark; Avg. averages all recalls.}
\label{tab:component_ablation}
\small
\setlength{\tabcolsep}{4.2pt}
\begin{tabular}{lcccc}
\toprule
Method & C-TL & A-TL & QV-TL & Avg. \\
\midrule
OPD            & 51.6 & 48.4 & 58.4 & 52.8 \\
Attention-Only & 51.8 & 48.6 & 59.3 & 53.3 \\
Token-Only     & 52.0 & 49.4 & 60.3 & 53.9 \\
CAPD           & \textbf{52.6} & \textbf{50.7} & \textbf{62.0} & \textbf{55.1} \\
\bottomrule
\end{tabular}
\end{table}

\begin{table}[t]
\centering
\caption{Parameter ablation on TimeLens. C-TL, A-TL, and QV-TL are the mean recalls on each benchmark; Avg. averages all nine recall metrics.}
\label{tab:parameter_ablation}
\small
\setlength{\tabcolsep}{3.2pt}
\begin{tabular}{cccccc}
\toprule
Parameter & Value & C-TL & A-TL & QV-TL & Avg. \\
\midrule
\multirow{4}{*}{\shortstack[c]{Temporal Group\\($G$)}}
& 2  & 52.2 & 49.8 & 60.7 & 54.2 \\
& 4  & 51.9 & 49.5 & 60.5 & 54.0 \\
& \textbf{8}  & \textbf{52.6} & \textbf{50.7} & \textbf{62.0} & \textbf{55.1} \\
& 16 & 52.0 & 49.5 & 60.4 & 53.9 \\
\midrule
\multirow{6}{*}{\shortstack[c]{Attention Weight\\($\lambda_{\mathrm{att}}$)}}
& 0    & 52.0 & 49.4 & 60.3 & 53.9 \\
& 0.1  & 52.4 & 49.6 & 60.9 & 54.3 \\
& \textbf{0.25} & \textbf{52.6} & \textbf{50.7} & \textbf{62.0} & \textbf{55.1} \\
& 0.5  & 52.0 & 49.6 & 60.4 & 54.0 \\
& 0.75 & 52.1 & 49.3 & 59.7 & 53.8 \\
& 1    & 51.9 & 49.0 & 59.5 & 53.5 \\
\bottomrule
\end{tabular}
\end{table}

\begin{figure*}[t!]
    \centering
    \includegraphics[width=\textwidth, trim=0cm 0cm 0cm 0cm]{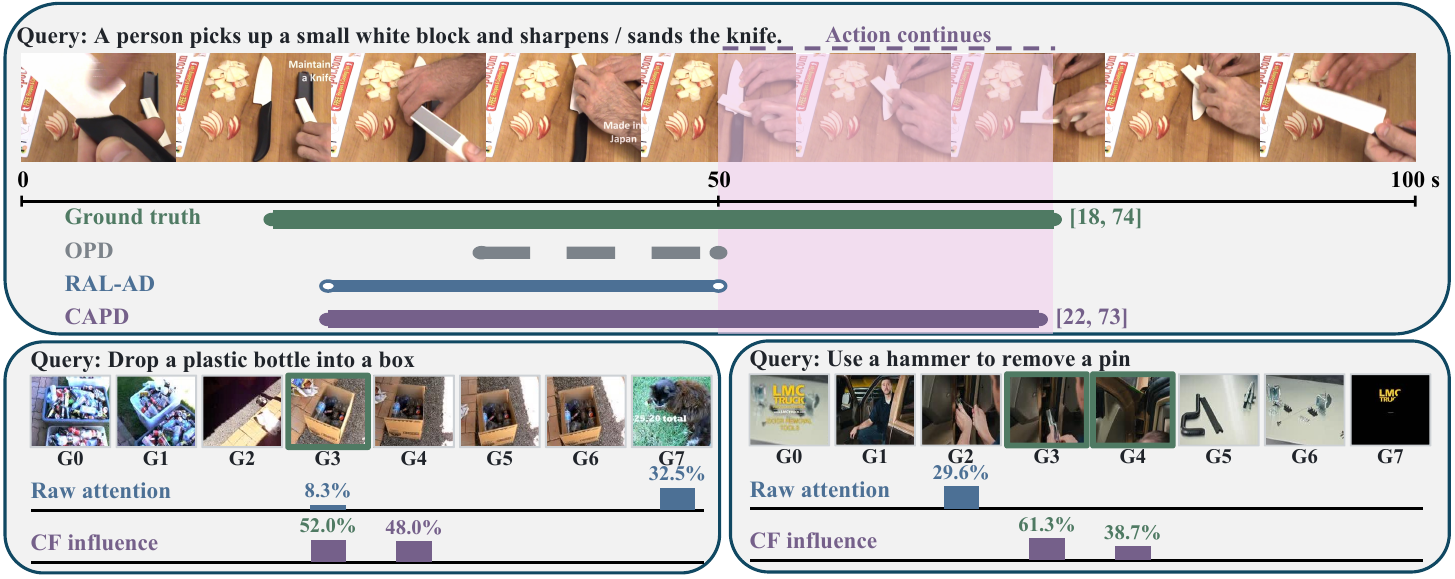}
    \caption{Visualized results of CAPD. The top panel shows the temporal boundary predictions of different methods on a long-form video query, alongside the ground-truth interval, showcasing CAPD's ability to produce compact and accurate temporal boundaries. The \textcolor[RGB]{74,160,74}{\textbf{GREEN}} bar denotes the ground truth, the \textcolor[RGB]{128,128,128}{\textbf{GRAY}} segments are OPD's predictions, the \textcolor[RGB]{70,130,180}{\textbf{BLUE}} line represents RAL-AD, and the \textcolor[RGB]{112,76,154}{\textbf{PURPLE}} bar indicates CAPD's output. The bottom-left and bottom-right panels contrast raw teacher attention with counterfactual (CF) influence across temporal groups G0--G7 for two additional queries, demonstrating that while raw attention concentrates on late or visually salient groups, CF influence more precisely identifies the relevant temporal segments.}
    \label{fig:qualitative}
\end{figure*}

In Table~\ref{tab:component_ablation}, we ablate the two components of CAPD, i.e., counterfactually calibrated attention supervision and counterfactual token weighting, with OPD as the reference. Both components improve over OPD, confirming that counterfactual influence provides useful supervision for both evidence selection and output generation. Token-Only produces a larger gain than Attention-Only, indicating that weighting decision-sensitive output tokens provides a stronger training signal. Attention-Only also improves performance, showing that aligning the student with the calibrated temporal evidence policy is beneficial, though insufficient alone. Combining both components achieves the best performance across all benchmarks, confirming their complementary roles, where token weighting strengthens supervision on decision-sensitive outputs while calibrated attention improves the temporal evidence used to generate them.

In Table~\ref{tab:parameter_ablation}, we study the effects of the number of temporal groups \(G\) and the attention-loss coefficient \(\lambda_{\mathrm{att}}\). The analysis in Figure~\ref{fig:ablation_1} reveals a clear trade-off in temporal grouping. Increasing \(G\) improves group precision by isolating finer temporal regions, but reduces evidence coverage, preference gap, and selection stability because a complete event may be divided across multiple groups. Coarser grouping better preserves complete and stable event evidence, although each group contains more irrelevant context. This trade-off is reflected in the downstream results, where performance does not improve monotonically with finer grouping and \(G=8\) achieves the best overall result. Meanwhile, performance first improves and then declines as \(\lambda_{\mathrm{att}}\) increases. A moderate attention weight provides useful evidence supervision while preserving the contribution of output-level distillation, whereas an overly large weight can disturb this balance. We therefore use \(G=8\) and \(\lambda_{\mathrm{att}}=0.25\), which provide the best balance between precise evidence selection, complete event coverage, and output-level supervision. 

Overall, the component and parameter studies confirm that counterfactual token weighting and calibrated attention provide complementary output- and evidence-level supervision.

\subsection{Qualitative Analysis}

To better understand how counterfactual calibration improves temporal grounding, we visualize both the predicted temporal boundaries and the corresponding temporal supervision signals in Figure~\ref{fig:qualitative}. In the upper panel, OPD mainly captures the visually salient middle of the target event, while RAL-AD recovers its onset but still terminates the prediction before the action is complete. The continuous video frames show that the queried action remains ongoing after both predictions have ended. In contrast, CAPD recovers the decision-relevant event tail and predicts a more complete interval that closely matches the ground-truth boundary. This comparison shows that attention-policy transfer improves evidence localization, while counterfactual calibration further corrects the boundary decision.

The lower panel compares the supervision signals. RAL-AD distills raw teacher attention, which can assign substantial mass to visually salient but irrelevant content. In the left example, attention focuses on the final scene, whereas counterfactual influence highlights the groups containing the queried action. A similar pattern appears in the right example, where attention peaks on an irrelevant group while counterfactual influence concentrates on the groups covering the actual tool-use action. These examples show that counterfactual influence better identifies temporal evidence that affects the teacher's prediction and therefore provides a more reliable signal for calibrating attention supervision.

Together, the two panels connect temporal evidence selection with boundary generation. Raw attention shows where the teacher attends, whereas counterfactual influence identifies the temporal groups whose removal changes its output. CAPD preserves the teacher's attention structure while increasing the target mass of influential groups, helping the student retain the evidence needed to predict complete event boundaries. This calibration more directly connects the selected temporal evidence to boundary generation. The resulting improvement in start and end timestamps is consistent with the larger gains at stricter IoU thresholds in Table~\ref{tab:main_results}, supporting the CAPD design.

\section{Conclusion}

In this paper, we introduce CAPD, an on-policy distillation method for temporal grounding in video MLLMs. CAPD estimates the counterfactual influence of temporal groups to calibrate teacher attention and output-level distillation. Experiments on TimeLens benchmarks demonstrate that CAPD improves temporal localization over OPD and RAL-AD while preserving general video understanding ability. These results show that CAPD provides more effective supervision for temporal grounding by integrating temporal evidence into on-policy distillation. Despite its promising results, CAPD uses fixed temporal groups, which may not fully capture events with diverse durations and temporal structures. We will explore event-adaptive grouping with finer partitions for short actions and coarser partitions for extended events. CAPD requires multiple teacher forward passes, increasing post-training cost over OPD. Since the teacher is frozen, batched inference can reduce this overhead.

\clearpage

\bibliography{references}


\end{document}